\documentclass[10pt, conference, compsocconf]{IEEEtran}
\ifCLASSINFOpdf
\else
\fi

\usepackage{graphicx}
\begin{document}
%
\title{Application of Pre-training Models in Named Entity Recognition}


\author{\IEEEauthorblockN{Yu Wang\textsuperscript{1}\textsuperscript{2},Yining Sun\textsuperscript{1}\textsuperscript{2},Zuchang Ma\textsuperscript{*}\textsuperscript{1}\textsuperscript{2},Lisheng Gao\textsuperscript{1}\textsuperscript{2},Yang Xu\textsuperscript{1},Ting Sun\textsuperscript{3}}
\IEEEauthorblockA{\textsuperscript{1}AnHui Province Key Laboratory of Medical Physics and Technology,\\
Institute of Intelligent Machines, Hefei Institutes of Physical Sciences, Chinese Academy of Sciences\\
\textsuperscript{2}University of Science and Technology of China\\
\textsuperscript{3}School of Nursing, Bengbu Medical College\\
\textsuperscript{*}Corresponding author, Email: zuchangma@163.com}
}


%


\maketitle

\begin{abstract}
Named Entity Recognition (NER) is a fundamental Natural Language Processing (NLP) task to extract entities from unstructured data. The previous methods for NER were based on machine learning or deep learning. Recently, pre-training models have significantly improved performance on multiple NLP tasks. In this paper, firstly, we introduce the architecture and pre-training tasks of four common pre-training models: BERT, ERNIE, ERNIE2.0-tiny, and RoBERTa. Then, we apply these pre-training models to a NER task by fine-tuning, and compare the effects of the different model architecture and pre-training tasks on the NER task. The experiment results showed that RoBERTa achieved state-of-the-art results on the MSRA-2006 dataset. 
\end{abstract}


%
\IEEEpeerreviewmaketitle

\section{Introduction}
Named Entity Recognition (NER) is a basic and important task in Natural Language Processing (NLP). It aims to recognize and classify named entities, such as person names and location names\cite{R IEEE ACCESS}. Extracting named entities from unstructured data can benefit many NLP tasks, for example Knowledge Graph (KG), Decision-making Support System (DSS), and Question Answering system. Researchers used rule-based and machine learning methods for the NER in the early years\cite{R M. Song}\cite{R Y. Zhao}. Recently, with the development of deep learning, deep neural networks have improved the performance of NER tasks\cite{R Z. Huang}\cite{R Y. Xia}. However, it may still be inefficient to use deep neural networks because the performance of these methods depends on the quality of labeled data in training sets while creating annotations for unstructured data is especially difficult\cite{R J. Lee}. Therefore, researchers hope to find an efficient method to extract semantic and syntactic knowledge from a large amount of unstructured data, which is also unlabeled. Then, apply the semantic and syntactic knowledge to improve the performance of NLP task effectively. 

Recent theoretical developments have revealed that word embeddings have shown to be effective for improving many NLP tasks. The Word2Vec and Glove models represent a word as a word embedding, where similar words have similar word embeddings\cite{R word embedding}. However, the Word2Vec and Glove models can not solve the problem of polysemy. Researchers have proposed some pre-training models, such as BERT, ERNIE, and RoBERTa, to learn contextualized word embeddings from unstructured text corpus\cite{R BERT}\cite{R ERNIE}\cite{R RoBERTa}. These models not only solve the problem of polysemy but also obtain more accurate word representations. Therefore, researchers pay more attention to how to apply these pre-training models to improve the performance of NLP tasks.

The purpose of this paper is to introduce the structure and pre-training tasks of four common pre-trained models (BERT, ERNIE, ERNIE2.0-tiny, RoBERTa), and how to apply these models to a NER task by fine-tuning. Moreover, we also conduct experiments on the MSRA-2006 dataset to test the effects of different pre-training models on the NER task, and discuss the reasons for these results from the model architecture and pre-training tasks respectively.

\begin{figure*}[t] 
    \centering
    \includegraphics[scale=0.6]{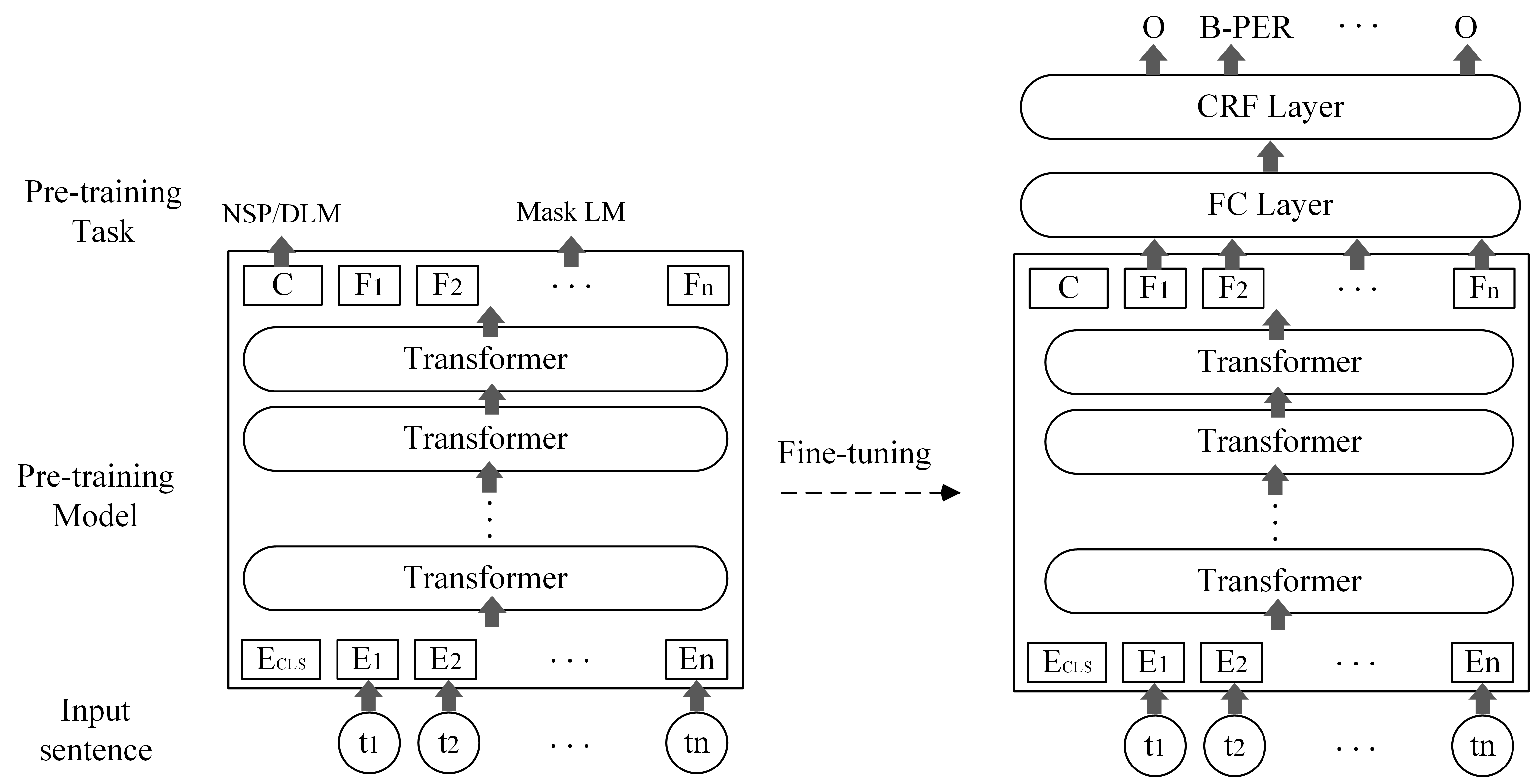}
    \caption{Fine-tuning}
    \label{Fig.fine-tune}
\end{figure*}
\section{Related work}

\subsection{Named Entity Recognition}
Named entity recognition (NER) is the basic task of the NLP, such as information extraction and data mining. The main goal of the NER is to extract entities (persons, places, organizations and so on) from unstructured documents. Researchers have used rule-based and dictionary-based methods for the NER\cite{R M. Song}. Because these methods have poor generalization properties, researchers have proposed machine learning methods, such as Hidden Markov Model (HMM) and Conditional Random Field (CRF)\cite{R Y. Zhao}\cite{R D. Li}. But machine learning methods require a lot of artificial features and can not avoid costly feature engineering. In recent years, deep learning, which is driven by artificial intelligence and cognitive computing, has been widely used in multiple NLP fields. Huang $et$ $al$. \cite{R Z. Huang} proposed a model that combine the  Bidirectional Long Short-Term Memory (BiLSTM) with the CRF. It can use both forward and backward input features to improve the performance of the NER task. Ma and Hovy \cite{R X. Ma} used a combination of the Convolutional Neural Networks (CNN) and the LSTM-CRF to recognize entities. Chiu and Nichols \cite{R J. Chiu} improved the BiLSTM-CNN model and tested it on the CoNLL-2003 corpus.

\subsection{Pre-training model}
As mentioned above, the performance of deep learning methods depends on the quality of labeled training sets. Therefore, researchers have proposed pre-training models to improve the performance of the NLP tasks through a large number of unlabeled data. Recent research on pre-training models has mainly focused on BERT. For example, R. Qiao $et$ $al$. and N. Li $et$ $al$. \cite{R R. Qiao}\cite{R N. Li} used BERT and ELMO respectively to improve the performance of entity recognition in chinese clinical records. E. Alsentzer $et$ $al$. , L. Yao $et$ $al$. and K. Huang $et$ $al$. \cite{R E. Alsentzer}\cite{R L. Yao}\cite{R K. Huang} used domain-specific corpus to train BERT(the model structure and pre-training tasks are unchanged), and used this model for a domain-specific task, obtaining the result of SOTA. 

\section{Methods}
In this section, we first introduce the four pre-trained models (BERT, ERNIE, ERNIE 2.0-tiny, RoBERTa), including their model structures and pre-training tasks. Then we introduce how to use them for the NER task through fine-tuning.

\subsection{BERT}
BERT is a pre-training model that learns the features of words from a large amount of corpus through unsupervised learning{\cite{R  BERT}}.

There are different kinds of structures of BERT models. We chose the \textbf{BERT-base} model structure. \textbf{BERT-base}'s architecture is a multi-layer bidirectional Transformer{\cite{R Attention}}. The number of layers is $L=12$, the hidden size is $H=768$, and the number of self-attention heads is $A=12${\cite{R BERT}}.

Unlike ELMO, BERT's pre-training tasks are not some kind of N-gram language model prediction tasks, but the "Masked LM (MLM)" and "Next Sentence Prediction (NSP)" tasks. For MLM, like a $Cloze$ task, the model mask 15\% of all tokens in each input sequence at random, and predict the masked token. For NSP, the input sequences are sentence pairs segmented with \texttt{[SEQ]}. Among them, only 50\% of the sentence pairs are positive samples.

\subsection{ERNIE}
ERNIE is also a pre-training language model. In addition to a basic-level masking strategy,  unlike BERT, ERNIE using entity-level and phrase-level masking strategies to obtain the language representations enhanced by knowledge {\cite{R ERNIE}}.

ERNIE has the same model structure as \textbf{BERT-base}, which uses 12 Transformer encoder layers, 768 hidden units and 12 attention heads.

As mentioned above, ERNIE using three masking strategies: basic-level masking, phrase-level masking, and entity-level masking. the basic-level making is to mask a character and train the model to predict it. Phrase-level and entity-level masking are to mask a phrase or an entity and predict the masking part. In addition, ERNIE also performs the "Dialogue Language Model (DLM)" task to judge whether a multi-turn conversation is real or fake {\cite{R ERNIE}}.

\subsection{ERNIE2.0-tiny}

ERNIE2.0 is a continual pre-training framework. It could incrementally build and train a large variety of pre-training tasks through continual multi-task learning \cite{R ERNIE2.0}.

ERNIE2.0-tiny compresses ERNIE 2.0 through the method of structure compression and model distillation. The number of Transformer layers is reduced from 12 to 3, and the number of hidden units is increased from 768 to 1024.

ERNIE2.0-tiny's pre-training task is called continual pre-training. The process of continual pre-training including continually constructing unsupervised pre-training tasks with big data and updating the model via multi-task learning. These tasks include word-aware tasks, structure-aware tasks, and semantic-aware tasks.

\subsection{RoBERTa}
RoBERTa is similar to BERT, except that it changes the masking strategy and removes the NSP task\cite{R RoBERTa}.

Like ERNIE, RoBERTa has the same model structure as BERT, with 12 Transformer layers, 768 hidden units, and 12 self-attention heads. 

RoBERTa removes the NSP task in BERT and changes the masking strategy from static to dynamic{\cite{R RoBERTa}}. BERT performs masking once during data processing, resulting in a single static mask. However, RoBoERTa changes masking position in every epoch. Therefore, the pre-training model will gradually adapt to different masking strategies and learn different language representations.

\subsection{Applying Pre-training Models}

After the pre-training process, pre-training models obtain abundant semantic knowledge from unlabeled pre-training corpus through unsupervised learning. Then, we use the fine-tuning approach to apply pre-training models in downstream tasks. As shown in Figure 1, we add the Fully Connection (FC) layer and the CRF layer after the output of pre-training models. The vectors output by pre-training models can be regarded as the representations of input sentences. Therefore, we use a fully connection layer to obtain the higher level and more abstract representations. The tags of the output sequence have strong restrictions and dependencies. For example, "I-PER" must appear after "B-PER".  Conditional Random Field, as an undirected graphical model, can obtain dependencies between tags. We add the CRF layer to ensure the output order of tags.

\section{Experiments and Results}
We conducted experiments on Chinese NER datasets to demonstrate the effectiveness of the pre-training models specified in section III. For the dataset, we used the MSRA-2006 published by Microsoft Research Asia.

The experiments were conducted on the AI Studio platform launched by the Baidu. This platform has a build-in deep learning framework PaddlePaddle and is equipped with a V100 GPU. The pre-training models mentioned above were downloaded by PaddleHub, which is a pre-training model management toolkit. It is also launched by the Baidu.  For hyper-parameter configuration, we adjusted them according to the performance on development sets. In this article, the number of the epoch is 2, the learning rate is 5e-5, and the batch size is 16.

\begin{table}[htbp]
	\centering  
	\caption{the results of NER using different pre-training models}  
	\label{table1}  
	\begin{tabular}{cccc}  
	    \hline  
		& & & \\[-6pt]  
		Models&Precision/\%&Recall/\%&F1/\%\\
		\hline  
		& & & \\[-6pt]  
		Baseline&92.54&88.20&90.32\\
		\hline
		& & & \\[-6pt]  
		BERT-base&92.68&94.18&93.30 \\
		\hline
		& & & \\[-6pt]  
	    ERNIE&92.92&94.07&93.37 \\
		\hline
		& & & \\[-6pt]  
	    ERNIE-tiny&83.89&89.88&86.52 \\
		\hline
		& & & \\[-6pt]  
	    RoBERTa&\textbf{93.64}&\textbf{94.93}&\textbf{94.17} \\
		\hline
	\end{tabular}
\end{table}

The BiGRU+CRF model was used as the baseline model. Table I shows that the baseline model has already achieved an F1 value of 90.32. However, using the pre-training models can significantly increase F1 values by 1 to 2 percentage points except for ERNIE-tiny model. Among the pre-training models, the RoBERTa model achieves the highest F1 value of 94.17, while the value of ERNIE-tiny is relatively low, even 4 percentage points lower than the baseline model.

\section{Discussion}

This section discusses the experimental results in detail. We will analyze the different model structures and pre-training tasks on the effect of the NER task. 

First of all, it is shown that the deeper the layer, the better the performance. All pre-training models have 12 Transformer layers, except ERNIE2.0-tiny. Although Ernie2.0-tiny increases the number of hidden units and improves the pre-training task with continual pre-training, 3 Transformer layers can not extract semantic knowledge well. The F1 value of ERNIE-2.0-tiny is even lower than the baseline model.

Secondly, for pre-training models with the same model structure, RoBERTa obtained the result of SOTA. BERT and ERNIE retain the sentence pre-training tasks of NSP and DLM respectively, while RoBERTa removes the sentence-level pre-training task because Liu $et$ $al$. \cite{R RoBERTa} hypothesizes the model can not learn long-range dependencies. The results confirm the above hypothesis. For the NER task, sentence-level pre-training tasks do not improve performance. In contrast, RoBERTa removes the NSP task and improves the performance of entity recognition. As described by Liu $et$ $al$. \cite{R RoBERTa}, the NSP and the MLP are designed to improve the performance on specific downstream tasks, such as the SQuAD 1.1, which requires reasoning about the relationships between pairs of sentences. However, the results show that the NER task does not rely on sentence-level knowledge, and using sentence-level pre-training tasks hurts performance because the pre-training models may not able to learn long-range dependencies.

Moreover, as mentioned before, RoBERTa could adapt to different masking strategies and acquires richer semantic representations with the dynamic masking strategy. In contrast, BERT and ERNIE use the static masking strategy in every epoch. In addition, the results in this paper show that the F1 value of ERNIE is slightly lower than BERT. We infer that ERNIE may introduce segmentation errors when performing entity-level and phrase-level masking.

\section{Conclusion}
In this paper, we exploit four pre-training models (BERT, ERNIE, ERNIE2.0-tiny, RoBERTa) for the NER task. Firstly, we introduce the architecture and pre-training tasks of these pre-training models. Then, we apply the pre-training models to the target task through a fine-tuning approach. During fine-tuning, we add a fully connection layer and a CRF layer after the output of pre-training models. Results showed that using the pre-training models significantly improved the performance of recognition. Moreover, results provided a basis that the structure and pre-training tasks in RoBERTa model are more suitable for NER tasks.

In future work, investigating the model structure of different downstream tasks might prove important.

\section*{Acknowledgment}
This research was funded by the major special project of Anhui Science and Technology Department (Grant: 18030801133) and Science and Technology Service Network Initiative (Grant: KFJ-STS-ZDTP-079).



%

\end{document}